\documentclass[11pt]{article}

\usepackage{ACL2023}

\usepackage{times}
\usepackage{latexsym}

\usepackage[T1]{fontenc}

\usepackage[utf8]{inputenc}

\usepackage{microtype}

\usepackage{inconsolata}

\usepackage{glossaries}
\usepackage{booktabs}
\usepackage{xcolor}
\usepackage{tikz}
\usepackage{adjustbox}
\usepackage{pifont}
\usepackage{amssymb}
\usepackage{hyperref}
\usepackage{enumitem}
\usepackage{verbatim}

\newacronym{nlp}{NLP}{Natural Language Processing}
\newacronym{dfm}{DFM}{Danish Foundation Models}
\newacronym{dagw}{DAGW}{Danish Gigaword}
\newacronym{dcc}{DCC}{Danish Colossal Corpus}
\newacronym{nat}{NAT}{Netarkivet Text}
\newacronym{seb}{SEB}{Scandinavian Embedding Benchmark}
\newacronym{asr}{ASR}{Automatic Speech Recognition}

\newcommand*\numcircledtikz[1]{\tikz[baseline=(char.base)]{
            \node[shape=circle,draw,inner sep=1.2pt] (char) {#1};}} 

\definecolor{ForestGreen}{RGB}{34,139,34}
\newcommand{\greencheck}{{\color{ForestGreen}\vspace{-0.3cm}\ding{51}\vspace{-0.3cm}}}
\newcommand{\greycross}{{\color{gray}\vspace{-0.3cm}\ding{55}\vspace{-0.3cm}}}

\title{Danish Foundation Models}

\author{
Kenneth Enevoldsen\Thanks{Equal Contributions} \textsuperscript{1,2} \And 
Lasse Hansen$^*$\textsuperscript{2,1} \And 
Dan S. Nielsen\textsuperscript{3} \AND 
Rasmus A. F. Egebæk\textsuperscript{4} \And 
Søren V. Holm\textsuperscript{4} \And 
Martin C. Nielsen\textsuperscript{4} \AND 
Martin Bernstorff\textsuperscript{2, 1} \And 
Rasmus Larsen\textsuperscript{3} \And 
Peter B. Jørgensen\textsuperscript{3} \AND
Malte Højmark-Bertelsen\textsuperscript{5} \And 
Peter B. Vahlstrup\textsuperscript{1} \And 
Per Møldrup-Dalum\textsuperscript{1} \AND
Kristoffer Nielbo\textsuperscript{1}
\\
\\
\footnotesize \textsuperscript{1}Center for Humanities Computing, Aarhus University, Denmark \\
\footnotesize \textsuperscript{2}Department of Clinical Medicine, Aarhus University, Denmark \\
\footnotesize \textsuperscript{3}The Alexandra Institute, Copenhagen, Denmark \\
\footnotesize \textsuperscript{4}Alvenir, Copenhagen, Denmark \\
\footnotesize \textsuperscript{5}Beyond Work \\ 
\footnotesize \texttt{kenneth.enevoldsen@cas.au.dk} \\
\footnotesize \texttt{lasse.hansen@clin.au.dk} \\
}

\begin{document}
\maketitle

\begin{abstract}
    Large language models, sometimes referred to as foundation models, have transformed multiple fields of research. However, smaller languages risk falling behind due to high training costs and small incentives for large companies to train these models. To combat this, the Danish Foundation Models project seeks to provide and maintain open, well-documented, and high-quality foundation models for the Danish language. This is achieved through broad cooperation with public and private institutions, to ensure high data quality and applicability of the trained models. We present the motivation of the project, the current status, and future perspectives. 
\end{abstract}

\section{Introduction}

In recent years, the field of machine learning has witnessed a paradigm shift, driven by the emergence of \emph{foundation models}: Models that are pre-trained on large quantities of data, that can be adapted to multiple downstream tasks \citep{bommasani_opportunities_2021, devlin_bert:_2019}. 
Training larger models on more extensive and diverse datasets has demonstrated improved performance across tasks \citep{brown_language_2020, kaplan_scaling_2020, touvron_llama_2023}. 
As a consequence, training better foundation models has become an active area of research. Strategies for this include increasing computational resources, increasing dataset sizes \citep{kaplan_scaling_2020}, and improving training efficiency by, e.g., removing low-quality data samples \citep{rae_scaling_2021}, proposing new training regimes \citep{clark2020electra}, new model architectures \citep{sun_retentive_2023}, or changes to, e.g., the attention mechanism \citep{dao2023flashattention}.

\begin{table*}[h]
\centering
\caption{A representative sample of different foundation model categories used for Danish: structured learning (e.g. encoders), generative models (e.g. decoders), search and ranking models (embeddings), and speech. Languages are denoted using a flag and multilingual is denoted with a globe.}
\begin{adjustbox}{width=0.9\textwidth}
\includegraphics[]{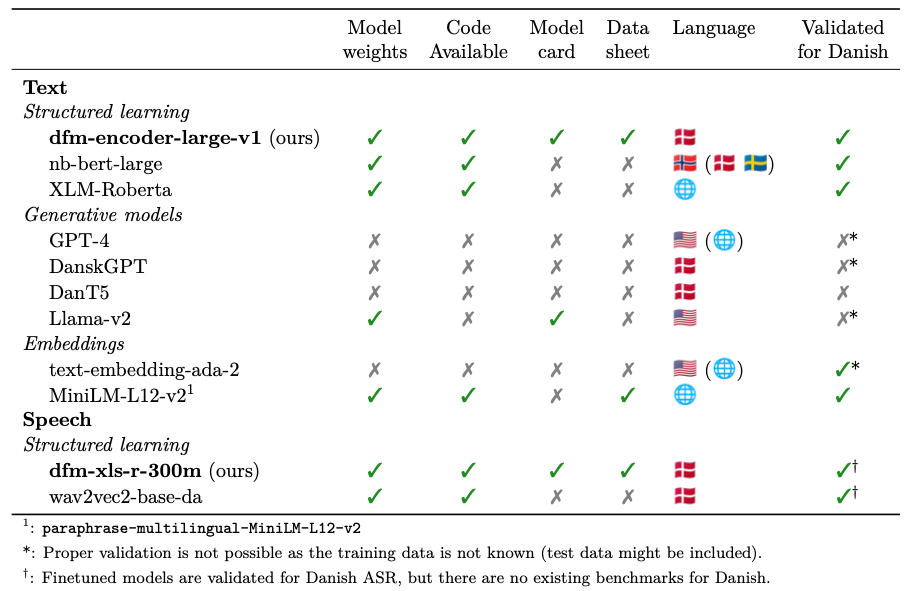}
\end{adjustbox}
\end{table*}

\subsection{The Case for Danish Foundation Models}

Foundation models are predominantly developed for the English language, with only a few models developed with multilingual capabilities (e.g. \cite{workshop_bloom_2022}). Although models trained on mostly English data have shown impressive performance on languages with limited representation in the training data \cite{zhu2023extrapolating}, these models inherently carry assumptions and cultural biases that may not seamlessly transfer between languages and cultures \cite{cao_assessing_2023}. For example, norms related to firearms or social security and welfare differ markedly between USA and Denmark. With respect to spoken language, Danish has a very distinct phonological structure \citep{basboll_phonology_2005, trecca_danish_2021}. 

Despite this, multilingual models perform well on specific benchmarks for low-resource languages, for instance, in \gls{seb} \citep{kenneth_enevoldsen_2023_10078412} -- which seeks to evaluate the document representations of a model -- multilingual models achieve superior performance for Danish and Swedish. However, evidence from high-quality benchmarks demonstrates that purpose-built monolingual models, or those restricted to closely related languages, often outperform their multilingual counterparts. Examples of this can be seen in the success of Norwegian, Danish, and Swedish models as documented in ScandEval \citep{nielsen2021scandeval} or of Italian models in UINAUIL \cite{basile2023uinauil}.

Developing Danish foundation models by a public institution becomes imperative due to the limited incentive for large tech companies to invest in languages spoken by smaller populations. Additionally, foundation models applied nationally seek to solve a different set of tasks than general-purpose models. Use cases such as healthcare services or citizen-state interactions will have high priority, while assistive technologies, e.g., programming, will be less central for national models \cite{kommunernes_landsforening_sprogmodeller_2023, syddansk_sundhedsinnovation_potentialet_2023}. National use cases also require different restrictions relating to privacy and governance, which necessitate local solutions without the need to send sensitive data to foreign service providers.

In the Nordics, collaborations between academia and libraries can be particularly fruitful due to similar archival laws and the possibility for data-sharing agreements, which allow the sharing of extensive and otherwise closed-source, resources with the research community. This makes it possible to publish models trained on otherwise inaccessible, high-quality datasets\cite{kummervold_operationalizing_2021}.

The field of Danish NLP is steadily growing, with datasets for pre-training \citep{stromberg-derczynski_danish_2021}, multiple task-specific datasets \citep{zeinert_annotating_2021, dansk}, and several pre-trained and fine-tuned models \citep{hojmark-bertelsen_aelaectra_2021,mollerhoj_danish_2019, ciosici_training_2022}. Despite this growth, the current landscape remains constrained in several key dimensions. First, the diversity of model architectures is severely limited, with no practically usable and open generative models. This significantly limits the potential applications within Danish contexts, as generative models are fundamental for developing tools such as virtual assistants. Second, prior to the \gls{dfm} project, a concerning trend highlighted by benchmark evaluations from ScandEval and \gls{seb}, was that multilingual -- and even monolingual Norwegian models -- outperformed their Danish counterparts \citep{nielsen2021scandeval, kenneth_enevoldsen_2023_10078412}. This performance disparity can be attributed to the following factors: \numcircledtikz{1} \emph{computational resources}: Existing Danish models have been trained for relatively few compute hours on modest hardware, compared to their international counterparts \citep{ciosici_training_2022, hojmark-bertelsen_aelaectra_2021, rae_scaling_2021}. \numcircledtikz{2} \emph{optimal resource utilization}: Most Danish language models use older architectures that are neither as compute- nor data-efficient as newer model architectures \citep{he_debertav3_2021,clark_electra_2020,devlin_bert:_2019}. \numcircledtikz{3} \emph{lack of training data}: models are trained on DAGW \citep{stromberg-derczynski_danish_2021} or the Danish part of Common Crawl, which has at least 100x fewer tokens than modern English models (e.g. \cite{rae_scaling_2021}). Similarly, prior to \gls{dfm}, Danish models have typically only employed minor filtering on the data source even though near-deduplication and quality filters have been shown to be beneficial \cite{rae_scaling_2021, lee_deduplicating_2021}. 

A crucial aspect of the current state of Danish language models is the mode of development and maintenance. Many of these models are created by motivated individuals who, unfortunately, lack the necessary resources and incentives to dedicate significant time to the documentation and maintenance of their models and datasets. This documentation, which includes aspects such as model cards and datasheets \citep{gebru_datasheets_2021, mitchell_model_2019}, is paramount for adoption in critical applications such as healthcare or public services. It provides valuable insights into the models' capabilities and highlights potential biases and limitations, a crucial step towards responsible and ethical AI development. Additionally, Danish lacks a number of benchmarks for evaluating the quality of language models. For example, no benchmarks exist for text generation or search and retrieval.

Although the Danish language model ecosystem is evolving, it faces critical challenges regarding model diversity, scale, data quality, and documentation. Addressing these limitations is essential for the Danish NLP community to build robust, effective, and responsible language models that can meet the unique linguistic and cultural nuances of the Danish language, as well as the Danish national context.

\section{Danish Foundation Models}

To resolve these issues, we present the \gls{dfm} project as a broad and open collaboration between academia, industry, and the open-source community. The \gls{dfm} project has four main aims:

\begin{enumerate}[noitemsep]
    \item To develop and maintain \emph{state-of-the-art language models for Danish} for applications within both text and speech.
    \item To extensively \emph{validate} foundation models for Danish in a representative set of tasks.
    \item To maintain a high standard of \emph{documentation} of models such as model cards \cite{mitchell_model_2019} and datasheets \cite{gebru_datasheets_2021}.
    \item To \emph{open-source} not only the models but also all components required for reproducibility such as pre-processing, training, and validation code.
\end{enumerate}

Through these four guiding principles, \gls{dfm} aspires to increase the quality and adoption of Danish language models. The emphasis on validating all relevant models enables users to critically assess and evaluate which model is best suited for their task, whether Danish, multilingual, or proprietary solutions. The high standard of documentation of \gls{dfm} models promises to provide trustworthy and transparent models in compliance with expected EU regulations. Datasets will be open to the extent possible within GDPR and proprietary restrictions.

\subsection{Dataset}

To bridge this gap, we present the \gls{dcc}: a composite corpus of text and speech from multiple domains. The text portion of \gls{dcc} consists of the Danish Gigaword Corpus \cite{stromberg-derczynski_danish_2021}, reddit-da, HopeTwitter\footnote{The situation with X.com (formerly Twitter) will be monitored closely with regard to any new restrictions or legal rulings that could impact this data collection.}, DaNews, and \gls{nat}. \gls{dcc} contains >100 billion text tokens spanning distinctly different domains, including news, social media, web, legal documents, and more. The speech portion consists of DaRadio, and DaTV, covering approximately 140.000 hours of unlabelled speech, and 900 hours of transcribed speech.

All subcorpora are extensively documented in datasheets \citep{gebru_datasheets_2021} that can be found on the project's \href{https://centre-for-humanities-computing.github.io/danish-foundation-models/}{website}. \gls{dcc} has been thoroughly pre-processed to ensure the highest data quality and scripts are available on the project repository. See Table~\ref {tab:datasets} for an overview of the contents of \gls{dcc}.

As the \gls{dcc} is composed of heterogeneous datasets, no common procedure for sharing the data can encompass the entirety of \gls{dcc}. Some of the data can be shared as-is and in the open (DAGW, reddit-da), others contain sensitive information on persons and can only be shared within the given laws and regulations, while others are the property of organizations that university partners have data processing and data transfer agreements with. For each sub-corpora of \gls{dcc}, a process for getting access within the given legislation is described in the relevant datasheets mentioned above.

\begin{table}
    \centering
    \caption{Overview of the subcorpora in \gls{dcc}.}
    \label{tab:datasets}
    \begin{adjustbox}{width=0.5\textwidth}
    \begin{tabular}[t]{lllp{1.2cm}p{1.2cm}}
        \toprule
        Name & Description & Size & Open access & Novel corpus\\
        \midrule
        \addlinespace[0.3em]
        \multicolumn{5}{l}{\textbf{Text}}\\
        \hspace{1em}DAGW & Danish Gigaword  & 1B tokens & \greencheck & \greycross\\
        \hspace{1em}reddit-da & Danish Reddit & <.1B tokens & \greencheck & \greycross\\
        \hspace{1em}HopeTwitter & Danish Tweets & 0.48B tokens & \greycross & \greencheck \\
        \hspace{1em}DaNews & Danish newspapers & 0.5B tokens & \greycross & \greencheck \\
        \hspace{1em}Netarkivet Text & Danish internet & >100B tokens & \greycross & \greencheck\\
        \addlinespace[0.3em]
        \multicolumn{5}{l}{\textbf{Speech}}\\
        \hspace{1em}DaRadio & Danish talk radio & 140.000 hours & \greycross & \greencheck\\
        \hspace{1em}DaTV & Danish subtitled TV & 900 & \greycross & \greencheck \\
        \bottomrule
    \end{tabular}
    \end{adjustbox}
\end{table}

\subsection{Current Achievements}

Currently, \gls{dfm} has created state-of-the-art models for speech and text processing and developed a benchmark for Scandinavian embedding models. The best performing Danish text model is \texttt{dfm-encoder-large-v1}, according to the ScandEval benchmark \citep{nielsen_scandeval_2023}. \texttt{dfm-encoder-large-v1} is a continued pre-training of an existing Scandinavian encoder model\footnote{\texttt{NbAiLab/nb-bert-large}} trained on the text portion of the \gls{dcc}. \texttt{dfm-encoder-large-v1} has already been integrated into tools such as DaCy \citep{enevoldsen_dacy_2021} and is used within both research and the private and public sectors\footnote{Personal communication.}.

For speech, a continued pre-training of XLS-R \cite{babu_xls-r_2021} for 120,000 steps on DaRadio has been released as \texttt{xls-r-300m-danish}. Only one other model of this type exists for Danish, however, it has 95M parameters compared to 300M in \texttt{xls-r-300m-danish}, is trained on a vastly smaller dataset, and does not perform as well. A fine-tuned version of the Danish XLS-R model, \texttt{xls-r-300m-danish-nst-cv9}, is currently the best-performing wav2vec-based \gls{asr} model for Danish.

Lastly, a new benchmark \gls{seb} has been created to investigate and compare embedding models across a wide range of tasks. \gls{seb} seeks to evaluate the document representation of existing models and thus does not fine-tune the full model. Instead, it utilizes the document embeddings for classification, bitext-mining, classification, and ranking. The benchmark is planned to expand to address the under-representation of ranking and search within existing Danish benchmarks.

All models are publicly available on the Hugging Face Hub under the \href{https://huggingface.co/chcaa}{Center for Humanities Computing (CHCAA)} organization.

\subsection{Planned Projects}

\gls{dfm} will continue to expand on multiple parallel tracks. On the model track, the current priority is to train and open-source generative text models as no such open model ready for production currently exists for the Danish language. Funding has been secured for training a 7B parameter model as a proof-of-concept, which will be further increased and developed in future iterations. In addition, we will fine-tune small and medium-sized Whisper models to improve the utility of \gls{asr} systems for the Danish language. 

Simultaneously, we will develop open benchmark tasks and datasets for generative models, to ensure and evaluate the quality of Danish and multi-lingual generative models. No such benchmark currently exists, which poses a significant barrier to developing and comparing different models. The Danish generative benchmark is set to include data from multiple domains such as healthcare, legal, question-answering, and more. The benchmark data and source code will be made publicly available to the extent possible.

\subsection{Future Perspectives}

The \gls{dfm} project is set to follow an iterative cycle of model and benchmark development, validation, refinement, and releases. We invite contributions and collaborators from industry and the open-source community alike.









\section*{Acknowledgements}

This research was supported by the ``HOPE - How Democracies Cope with COVID-19'' project funded by the Carlsberg Foundation with grant CF20-0044, DeiC Type-1 HPC with projects DeiC-AU1-L-000001, DeiC-AU-N1-000011, DeiC-AU-N1-000012, DeiC-AU-N5-2023034.

\newpage
\bibliography{refs}
\bibliographystyle{acl_natbib}

\end{document}